\documentclass[runningheads,a4paper]{llncs}

\usepackage{amssymb}
\usepackage{amssymb,amsmath,array}

\DeclareMathOperator*{\argmin}{arg\,min}

\usepackage{tikz}
\usetikzlibrary{arrows}

\usepackage{times}

\usepackage{graphicx}

\usepackage{subfig}
\usepackage[graphicx]{realboxes}

\usepackage{caption}
\usepackage{bbm}
\usepackage{datetime}
\usepackage[colorinlistoftodos]{todonotes}
\usepackage{color}
\definecolor{dblue}{rgb}{0,0,0.6}
\usepackage{amssymb}
\usepackage{relsize}
\usepackage{pifont}
\usepackage{multirow}

\usepackage{graphicx}

\usepackage{url}
\urldef{\mailsa}\path|{y.bahroun, a.soltoggio}@lboro.ac.uk|    
\newcommand{\keywords}[1]{\par\addvspace\baselineskip
\noindent\keywordname\enspace\ignorespaces#1}

\begin{document}

\mainmatter  % start of an individual contribution

\title{Online Representation Learning with Single and \\ Multi-layer Hebbian Networks for Image Classification}

\titlerunning{Online Representation Learning with Hebbian Networks}

\author{Yanis Bahroun \and Andrea Soltoggio}
\authorrunning{Online Representation Learning with Hebbian Networks}
\institute{Loughborough University, Computer Science Department,\\
LE11 3TU, Leicestershire, United Kingdom\\
\mailsa}


\maketitle

\begin{abstract}
Unsupervised learning permits the development of algorithms that are able to adapt to a variety of different data sets using the same underlying rules thanks to the autonomous discovery of discriminating features during training. 
Recently, a new class of Hebbian-like and local unsupervised learning rules for neural networks have been developed that minimise a \emph{similarity matching} cost-function. 
These have been shown to perform sparse representation learning.  This study tests the effectiveness of one such learning rule for learning features from images. 
The rule implemented is derived from a nonnegative classical multidimensional scaling cost-function, and is applied to both single and multi-layer architectures. 
The features learned by the algorithm are then used as input to a SVM to test their effectiveness in classification  on the established CIFAR-10 image dataset. 
The algorithm performs well in comparison to other unsupervised learning algorithms and multi-layer networks, thus suggesting its validity in the design of a new class of compact, online learning networks. 
\end{abstract}

\section{Introduction}
%
%Synaptic plasticity stands as one of the main phenomena responsible for learning and memory. One mechanism of synaptic weight update is inspired by the Hebbian learning principle which strengthens the connections between two units when they are simultaneously active. This principle makes the network learn recurring patterns.
%
%
Biological synaptic plasticity is hypothesized to be one of the main phenomena responsible for human learning and memory. One mechanism of synaptic plasticity is inspired by the Hebbian learning principle which states that connections between two units, e.g., neurons, are strengthened when they are simultaneously activated. 
%
%In the context of machine learning, this is realized through the mechanism of weight update, which makes a network learn recurring patterns. 
In artificial neural networks, implementations of Hebbian plasticity are known to learn recurring patterns of activations. 
The use of extensions of this rule, such as Oja's rule \cite{oja1989neural} or the Generalized Hebbian rule, also called Sanger's rule \cite{sanger1989optimal}, have permitted the development of algorithms that have proved particularly efficient at tasks such as online dimensionality reduction. 
%
%Extensions of the Hebb's rule such as the Oja's rule \cite{oja1989neural} or the Generalized Hebbian rule, also called Sanger's rule \cite{sanger1989optimal}, proved particularly efficient at tasks such as online dimensionality reduction.
%
Two important properties of brain-inspired models, namely competitive learning \cite{rumelhart1985feature} and sparse coding \cite{olshausen1996emergence} can be performed using Hebbian and anti-Hebbian learning rules. %, which permit the algorithm to forget previously acquired features when they prove irrelevant.} 
Such properties can be achieved with inhibitory connections, which extend the capabilities of such learning rules beyond simple extraction of the principal component of input data. 
%
%Formulating the learning problem under a minimisation principle offers a rigorous framework \cite{arora2015simple} to study the learning dynamics of the network.
%
%
% Moreover, the way the cost-function proposed in \cite{pehlevan2014hebbian} is derived leads to corresponding network architectures.
The continuous and local update dynamics of Hebbian learning also make it suitable for learning from a continuous stream of data. Such an algorithm can take one image at a time with memory requirements that are independent of the number of samples. 

This study employs Hebbian/anti-Hebbian learning rules derived from a similarity matching cost-function \cite{pehlevan2014hebbian} and applies it to perform online unsupervised learning of features from multiple image datasets. 
%
% Formulating the learning problem under a minimisation principle offers a rigorous framework \cite{arora2015simple} to study the learning dynamics of the network.
%
%
% A recent model of Hebbian/anti-Hebbian neural network \cite{pehlevan2014hebbian} was reproduced here and applied for the first time 
 The rule proposed in \cite{pehlevan2014hebbian} is applied here for the first time to online features learning for image classification with single and multi-layer architectures. The quality of the features is assessed visually and by performing classification with a linear classifier working on the learned features. 
The simulations show that a simple single-layer Hebbian network can outperform more complex models such as Sparse Autoencoders (SAE) and Restricted Boltzmann machines (RBM) for image classifications tasks \cite{coates2011analysis}. 
When applied to multi-layer architectures, the rule learns additional features. %which lead to further improvements of the classification accuracy.
This study is the first of its kind to perform multi-layer sparse dictionary learning based on the similarity matching principle developed in \cite{pehlevan2014hebbian} and to apply it to image classification. 
%Unlike SAE or RBM, that perform an energy or reconstruction error minimisation, \textcolor{dblue}{the algorithm used in this paper allows the network to learn similarity structures between inputs, structures also witnessed in the neural activities of the human object vision pathway \cite{connolly2012three}.}
%
%Moreover, the learning principle based on the multidimensional scaling cost-function considers input similarities also strongly relates to the representational similarity analysis developed in \cite{kriegeskorte2008representational} which appears critical for understanding the IT cortex. 
%
%One further advantage of the algorithm is that it is fully unsupervised and does not require any semi labeling nor data-augmentation.
%
%
% 
% ****************************************************************************
% ****************************************************************************
% ****************************************************************************
% ****************************************************************************
% ****************************************************************************
% Background AREA
% ****************************************************************************
% ****************************************************************************
% ****************************************************************************
% ****************************************************************************
% ****************************************************************************

\section{Hebbian/anti-Hebbian Network Derived From a \textcolor{black}{Similarity Matching} Cost-Function}
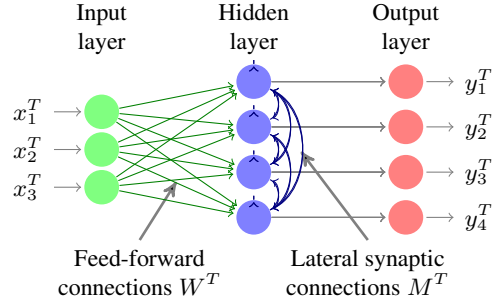
\begin{figure}[!ht]
\centering
\def\layersep{2cm}
\tikzstyle{information text}=[text badly centered,font=\small,text width=3cm]
\begin{tikzpicture}[shorten >=1pt,->,draw=black!50, node distance=\layersep]
    \tikzstyle{every pin edge}=[<-,shorten <=1pt]
    \tikzstyle{neuron}=[circle,fill=black!25,minimum size=13pt,inner sep=0pt]
    \tikzstyle{input neuron}=[neuron, fill=green!50];
    \tikzstyle{output neuron}=[neuron, fill=red!50];
    \tikzstyle{hidden neuron}=[neuron, fill=blue!50];
    \tikzstyle{annot} = [text width=4em, text centered]

 \begin{scope}[>=stealth', line width=1pt]
        \draw[->] (0.5,-2.2) node[below, information text]
            {Feed-forward connections $W^T$ } -- (1,-1.4);
        \draw[->] (3.5,-2.2) node[below,information text]
            {Lateral synaptic connections $M^T$} -- (2.6,-1);
    \end{scope}
    % Draw the input layer nodes
    \foreach \name / \y in {1,...,3}
    % This is the same as writing \foreach \name / \y in {1/1,2/2,3/3,4/4}
        \node[input neuron, pin=left:$x^T_\y$] (I-\name) at (0,-\y*0.5) {};

    % Draw the hidden layer nodes
    \foreach \name / \y in {1,...,4}
        \path[yshift=0.5cm]
            node[hidden neuron] (H-\name) at (\layersep,-\y*0.6 cm) {};

    % Draw the output layer node
    \foreach \name / \y in {1,...,4}
        \path[yshift=0.5cm]
            node[output neuron,pin={[pin edge={->}]right:$y^T_\y$}, right of=(O-\name)] (O-\name) at (\layersep,-\y*0.6 cm) {};    	

    % Connect every node in the input layer with every node in the
    % hidden layer.
    \foreach \source in {1,...,3}
        \foreach \dest in {1,...,4}
            \path (I-\source) edge[green!50!black] (H-\dest);

    % Connect every node in the hidden layer with the output layer
    \foreach \source in {1,...,4}
	    \foreach \dest in {1,...,4}
     	   \path (H-\source) edge (O-\source);

    % Connect every node in the hidden layer with itself layer
    \foreach \source in {1,...,4}
	    \foreach \dest in {\source,...,4}
     	   \path (H-\source)edge[blue!50!black,bend left=65] (H-\dest);
    \foreach \source in {1,...,4}
	    \foreach \dest in {\source,...,4}
		\path (H-\dest)edge[blue!50!black,bend right=65] (H-\source);

    % Annotate the layers
    \node[annot,above of=H-1, node distance=0.7cm] (hl) {Hidden layer};
    \node[annot,left of=hl] {Input layer};
    \node[annot,right of=hl] {Output layer};
\end{tikzpicture}
\caption{\label{fig:Network} Hebbian/anti-Hebbian network with lateral connections derived from Eq.\ref{eq:local}}
\end{figure}

The rule implemented by the Hebbian/anti-Hebbian network used in this work derives from an adaptation of Classical MultiDimensional Scaling (CMDS). CMDS is a popular embedding technique \cite{cox2000multidimensional}. Unlike most dimensionality reduction techniques, e.g. PCA, the CMDS uses as input the matrix of similarity between inputs to generate a set of embedding coordinates. The advantage of MDS is that any kind of distance or similarity matrix can be analyzed.
%It takes as input a  and generates a set of embedding 
%
However, in its simplest form, CMDS produces dense features maps which are often unsuitable when considered for image classification. 
% It is a fundamental information analysis tool with applications widely spread from ecology, eduction to neuroscience \cite{borg2005modern}.
%
%
% Dense features are also less biological plausible than sparse representations leading to higher classification accuracy  when linear classifier are used. 
% %
% A popular way of producing sparse codes is by imposing non-negativity constraints, which also ensure biological plausibility.
Therefore an adaptation of the CMDS introduced recently in \cite{pehlevan2014hebbian} is used to overcome this weakness. The model implemented is a nonnegative classical multidimensional scaling that has three properties: it takes a similarity matrix as input, it produces sparse codes, and can be implemented using a new biologically plausible Hebbian model.
%Recently, \cite{pehlevan2014hebbian} introduced a non-negative classical multidimensional scaling model which has three properties: it \textcolor{dblue}{takes a} similarity matrix as input, it produces sparse codes, and can be \textcolor{dblue}{implemented} using a new biologically plausible Hebbian model.
%
The Hebbian/anti-Hebbian rule introduced in \cite{pehlevan2014hebbian} is given as follows: for a set of inputs $x^t \in \mathbf{R}^n$ for $t \in \{1,\ldots,T\}$, the concatenation of the inputs defines an input matrix $X \in \mathbf{R}^{n \times T}$. The output matrix $Y$ of encodings is an element of $\mathbf{R}^{m \times T}$ that corresponds to a sparse overcomplete representation of the input if $m>n$, or to a low-dimensional embedding of the input if $m<n$.
The objective function proposed by \cite{pehlevan2014hebbian} is:
\begin{eqnarray}\label{eq:global}
Y^* = \argmin_{Y\geq 0} \| X'X -Y'Y \|_F^2 \quad 
\end{eqnarray}
where $F$ is the Frobenius norm and $X'X$ is the Gram matrix of the inputs which corresponds to the similarity matrix.
Solving Eq.\ref{eq:global} directly requires storing $Y\in \mathbf{R}_+^{m \times T}$ which increases with time $T$ making online learning difficult.
Thus instead an online learning version of Eq.\ref{eq:global} is expressed as:
\begin{eqnarray}\label{eq:local}
(y^T)^* = \argmin_{y^T \geq 0} \| X'X -Y'Y \|_F^2 \quad .
\end{eqnarray}
The components of the solution of Eq.\ref{eq:local}, found in \cite{pehlevan2014hebbian} using coordinate descent, are :
\begin{eqnarray}\label{activ_neurons}
(y^{T}_{i})^* = \max \bigg( W^{T}_{i}x^T -M^{T}_{i}(y^{T})^* ,0 \bigg) \quad \forall i \in \{1,\ldots, m\}, \quad 
\end{eqnarray}
\begin{eqnarray}
\text{where }\quad W^T_{ij}=\frac{\sum\limits_{t=1}^{T-1} y^t_{i}x^t_{j}}{\sum\limits_{t=1}^{T-1} (y^t_{i})^2} \quad ; \quad M^T_{ij} =\frac{\sum\limits_{t=1}^{T-1} y^t_{i}y^t_{j}}{\sum\limits_{t=1}^{T-1} (y^t_{i})^2}  \mathbf{1}_{i\neq j} \quad .
\end{eqnarray}
$W^T$ and $M^T$ can be found using the recursive formulations:
\begin{equation}\label{update_eq_hebb}
W^{T}_{ij}= W^{T-1}_{ij} + \bigg( y^{T-1}_{i}(x^{T-1}_{j} - W^{T-1}_{ij}y^{T-1}_{i})\bigg/ \hat Y^T_{i}\bigg)
\end{equation}
\begin{equation}\label{update_eq_anti}
M^{T}_{ij \neq i}= M^{T-1}_{ij} + \bigg(y^{T-1}_{i}(y^{T-1}_{j} - M^{T-1}_{ij}y^{T-1}_{i})\bigg/ \hat Y^T_{i} \bigg)
\end{equation}
\begin{equation}\label{update_eq_weight}
\hat Y^{T}_{i} =  \hat Y^{T-1}_{i} + (y^{T-1}_{i})^2 \quad .
\end{equation}
$W^T$ (green arrows) and $M^T$ (blue arrows) can be interpreted respectively as feed-forward synaptic connections between the input and the hidden layer and lateral synaptic inhibitory connections within the hidden layer. The weight matrices are of fixed sizes and updated sequentially, which makes the model suitable for online learning. The architecture of the Hebbian/anti-Hebbian network is represented in Figure \ref{fig:Network}.
\vspace{-8mm}
%
%
%
%
%The learning principle used in the following resembles a variant of the Hebbian principle presented in \cite{arora2015simple} performing alternating minimization.
%
% ****************************************************************************
% ****************************************************************************
% ****************************************************************************
% ****************************************************************************
% ****************************************************************************
% METHODOLOGY AREA
% ****************************************************************************
% ****************************************************************************
% ****************************************************************************
% ****************************************************************************
% ****************************************************************************
\section{A Model to Learn Features From Images}
In the new model presented in this study,  the input data vectors ($x^1,\ldots, x^T$) are composed of patches taken randomly from a training dataset of images. For every new input $x^t$ presented, the model first computes a sparse post-synaptic activity $y^t$. Second, the synaptic weights are modified based on local Hebbian/anti-Hebbian learning rules requiring only the current pre- post-synaptic neuronal activities.
The model can be seen as a sparse encoding followed by a recursive updating scheme, which are both well suited to solve large-scale online problems.
%
%An important aspect of this learning network is that it does not require batch learning, which in turn reduces the memory requirement of the learning system.
%
% A multi-class SVM classifies the pictures using the features learned by the neural network. A simple form of pooling is used to feed the classifier, the output vector is pooled over the four equal-sized quadrants of the image \cite{coates2011analysis} corresponding to $4 \times m$ dimensional feature vectors for each image.

A multi-class SVM classifies the pictures using output vectors obtained by a simple pooling of the feature vectors, $Y^*$, obtained for the input images from the trained network. In particular, given an input image, each neuron in the output layer produces a new image, called a feature map, which is pooled in quadrants \cite{coates2011analysis} to form 4 terms of the input vector for the SVM. 
\subsection{Multi-layer Hebbian/anti-Hebbian Neural Network}
In the proposed approach, layers of Hebbian/anti-Hebbian network are stacked similarly to the Convolutional DBN \cite{krizhevsky2010convolutional}, and Hierarchical K-means. %\cite{hu2012hierarchical} 
In the multi-layer Hebbian/anti-Hebbian network, both the weights of the first layer and second layer are continuously updated.
Unlike other CNNs, the non-linearity used in each layer is not only due to the positivity constraint, but to the combination of a rectified linear unit activation function and of interneuronal competition.
This model combines the powerful architecture of convolutional neural networks using ReLU activation with interneuronal competition, while all synaptic weights are updated using online local learning rules.
In between layers, a $2\times2$ average pooling is used to downsample the feature maps. %$\textcolor{dblue}{by a factor of 4}.
\subsection{Overcompleteness of the Representation and Multi-resolution}
As part of the evaluation of the new model, it is important to assess its performance with different sizes ($m$) of the hidden layers. 
%} proposed in this study, tests with different sizes of the hidden layers are important in order to evaluate the capabilities of the model.
%
If the number of neurons exceeds the size of the input ($m>n$), the representation is called overcomplete. Overcompleteness may be beneficial, but requires increased computation, particularly for deep networks in which the number of neurons has to grow exponentially in order to keep this property.
%in layer $N>2$ has to be bigger than the product of the size of a feature patch and the number of neurons in layer $N-1$.
%
One motivation for overcompleteness is that it may allow more flexibility in matching the output structure with the input.
However, not all learning algorithms can learn and take advantage of overcomplete representations. %\textcolor{black}{Overcompleteness is often} a characteristic shared by models performing sparse coding.
The behaviour of the algorithm is analysed in the transition between undercomplete ($m<n$) and overcomplete ($m>n$) representations.

%\subsection{Multi-resolution Hebbian/anti-Hebbian network}
%
Although the model might benefit from a large number of neurons, from a practical perspective an increase in the number of neurons is a challenge for such models due to the number of operations required in the coordinate descent. %The \textcolor{dblue}{practical} advantages of the algorithm are still \textcolor{dblue}{reduced} by the number of operations required by the coordinate descent when the number of neurons increases. 
In order to limit the computational cost of training a large network while still benefiting from overcomplete representations, this study proposes to train simultaneously three single-layer neural networks, each of them having different receptive field sizes ($4\times4, 6\times6,$ and $8\times8$ pixels). Thus, a variation of the model tested here is composed of three different networks. This architecture of parallel networks with different receptive field sizes requires less computational time and memory than a model with only one receptive field size and the same total number of neurons, because the synaptic weights only connect neurons within each neural network.
This model will be called multi-resolution in the following. %It can be seen as a biologically plausible, fully unsupervised and online inception single-layer network proposed in the GoogLeNet \cite{szegedy2015going}}.
\subsection{Parameters and Preprocessing}
The architecture used here has the following tunable parameters: the receptive field size ($n$) of the neurons and the number of neurons ($m$). 
These parameters are standard to CNNs but their influence on this online feed-forward model needs to be investigated. 

% However, since the network minimizes the nonnegative MDS cost-function, the influence of such parameters is unknown and needs to be evaluated.} 
%
%Unlike with K-means \cite{coates2011analysis} or CNNs, the algorithm does not use a cross-validated fixed threshold function reducing the number of tunable parameters of the model.
%

For computer vision models, understanding the influence of input preprocessing is of critical importance for both biological plausibility and practical applicability. Recent findings \cite{abbasi2016retinal}, confirm partial decorrelation of the input signal in the retinal ganglion cells. The influence of input decorrelation by applying whitening will be investigated.
%on the learning capabilities of the model 

% ****************************************************************************
% ****************************************************************************
% ****************************************************************************
% ****************************************************************************
% ****************************************************************************
% RESULTS AREA
% ****************************************************************************
% ****************************************************************************
% ****************************************************************************
% ****************************************************************************
% ****************************************************************************

\section{Results}

The effectiveness of the algorithm is assessed by measuring the performance on an image classification task. 
We acknowledge that classification accuracy is at best an implicit measure evaluating the performance of representation learning algorithms, but provides a standardised way of comparing them. In the following, single and multi-layer Hebbian/anti-Hebbian neural networks combined with the standard multi-class SVM are trained on the CIFAR-10 dataset \cite{krizhevsky2009learning}.
\subsection{Evaluation of the Single-layer Model}
\begin{figure}[!ht]
  \centering
  \label{figur}\caption{Sample of features learned from raw (\ref{figur:cent_unwhiten}) and whitened input (\ref{figur:cent_whiten}). Classification accuracy with raw (\ref{figur:rf_unwhit_Hebb}) and whitened input (\ref{figur:rf_whit_Hebb}).}
  \subfloat[Features learned from raw data]{\label{figur:cent_unwhiten}\includegraphics[width=50mm]{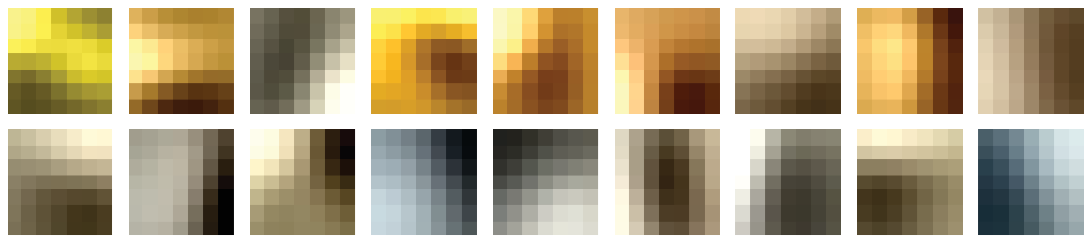}} \quad \quad
    \subfloat[Features learned from whitened data]{\label{figur:cent_whiten}\includegraphics[width=50mm]{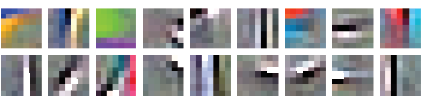}}
\\
\subfloat[Accuracy using raw data]{\label{figur:rf_unwhit_Hebb}\includegraphics[width=45mm]{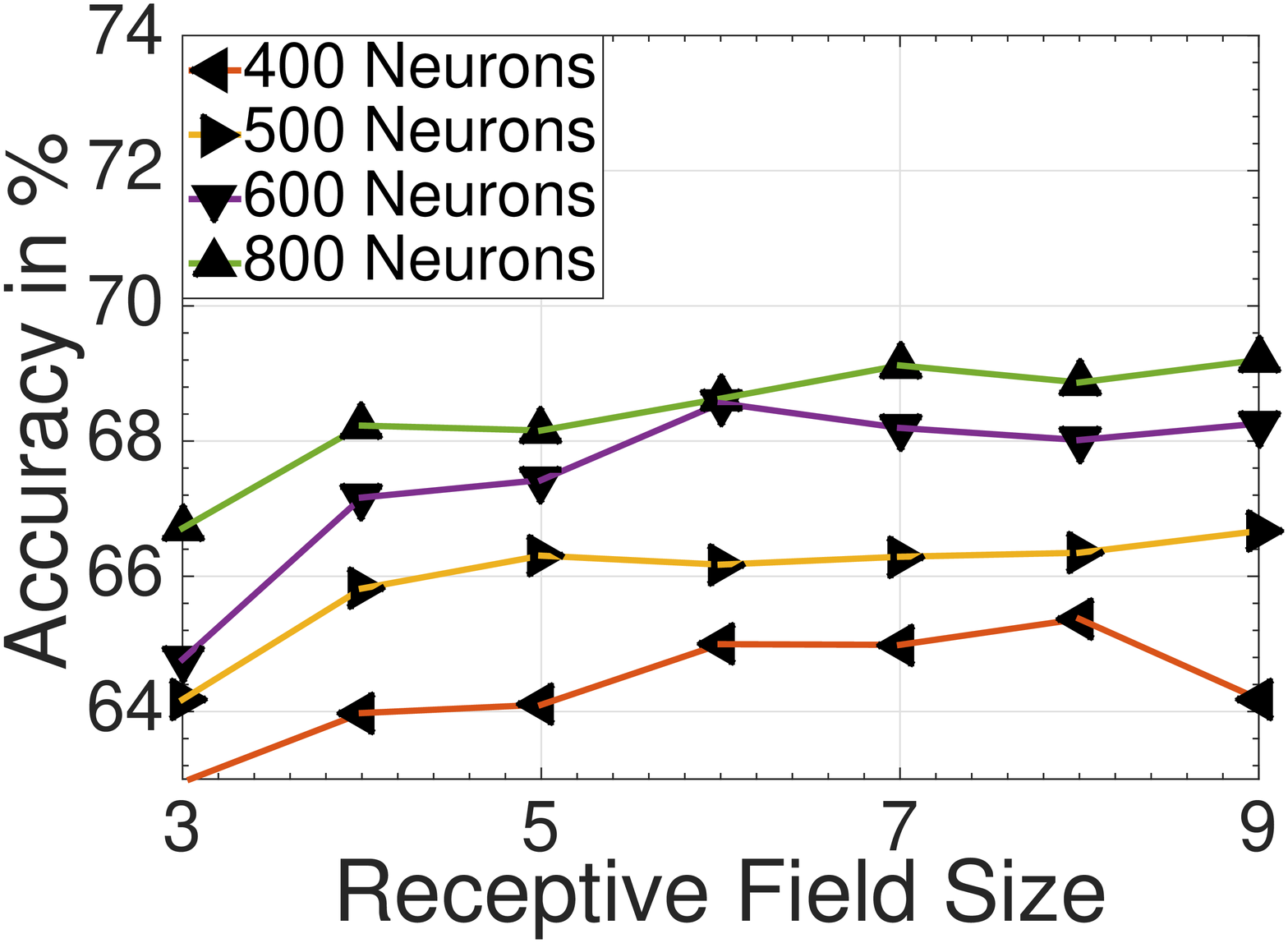}}
 \quad \quad \quad    \subfloat[Accuracy using whitened data]{\label{figur:rf_whit_Hebb}\includegraphics[width=45mm]{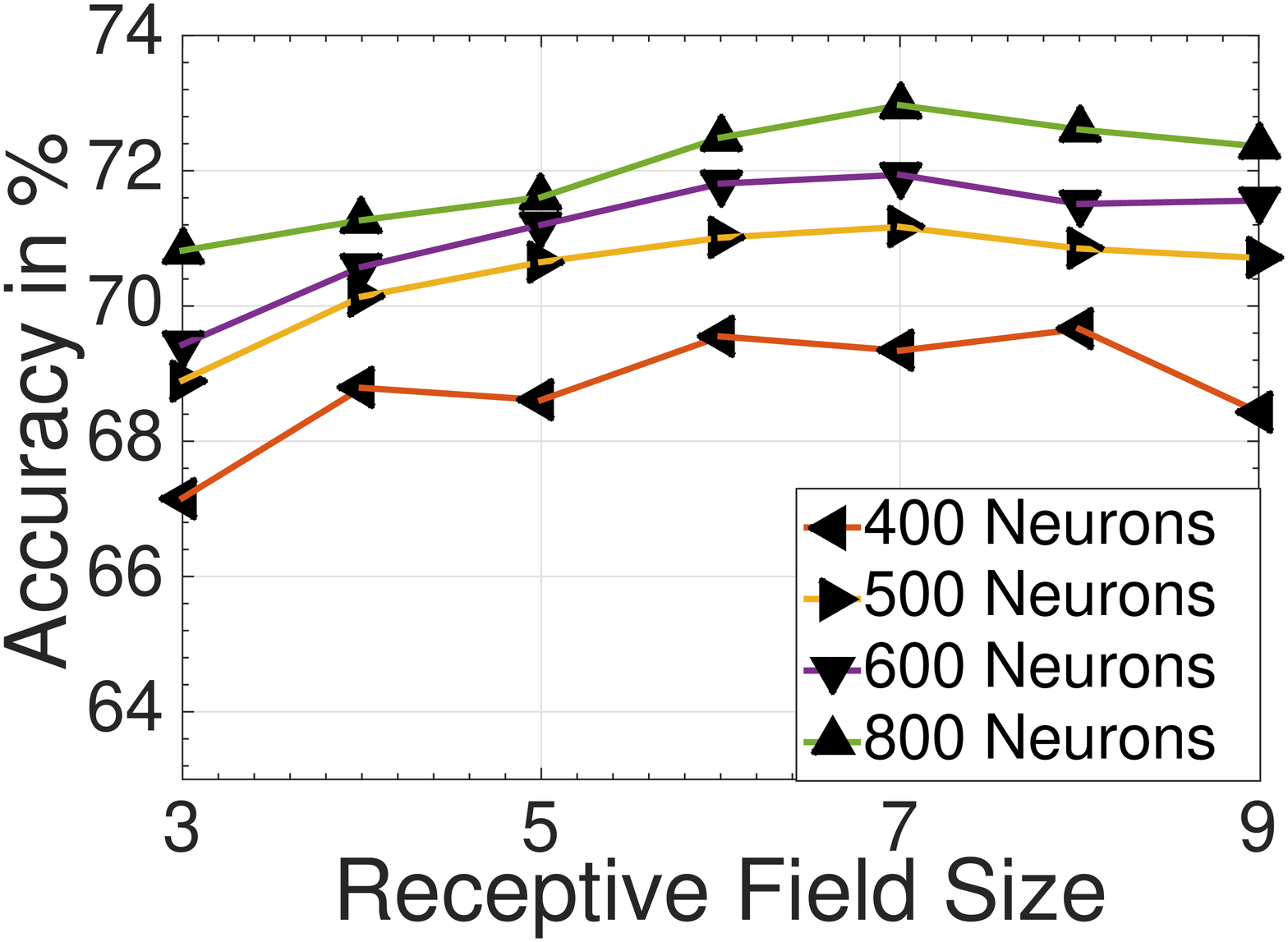}}
\end{figure}

A first experiment tested the performance of the model with and without whitening of the input data.
%
%In the following, brightness and contrast normalisation were applied to the input images.
Although there exist Hebbian networks that can perform online whitening \cite{pehlevan2015normative}, an offline technique based on singular value decomposition \cite{coates2011analysis} is applied in these experiments.
Figure~\ref{figur:cent_unwhiten} and ~\ref{figur:cent_whiten} show the features learned by the network from raw input and whitened input respectively. 
The features learned from raw data (Fig.\ref{figur:cent_unwhiten}) are neither sharp nor localised filters and just slightly capture edges. With whitened data (Fig.\ref{figur:cent_whiten}), the features are sharp, localised, and resemble Gabor filters, which are observed in the primary visual cortex \cite{olshausen1996emergence}.
%
%These results match those reported with clustering algorithms \cite{coates2011analysis}.
%
%
%\vspace{-5mm}

In a second set of experiments, the performance of the network was tested for varying receptive field sizes (Fig.\ref{figur:rf_unwhit_Hebb}-\ref{figur:rf_whit_Hebb}) and varying network sizes (400, 500, 600, and 800 neurons). The results show that the performance peaks at a receptive field size of 7 pixels and then begins to decline. 
% since no neural network minimizing such cost-function have been tested in this context
This property is common to most unsupervised learning algorithms \cite{coates2011analysis}, showing the difficulty of learning spatially extended features.
Figures~\ref{figur:rf_unwhit_Hebb} and \ref{figur:rf_whit_Hebb} also show that for every configuration, the performance of the algorithm is largely and uniformly improved when whitening is applied to the input.
%
%Finally, the step-size between extracted features (stride) is a meta-parameter  influencing the classification accuracy. Preliminary experiments (not shown) indicated that a stride of 1 was optimal, which is used for all the experiments presented in this work.

\subsection{Comparison to State-of-the-art Performances and Online Training}
\begin{figure}[!ht]
  \centering
  \label{fig:Classif2}\caption{(a) Proposed model vs K-means, (b) Classification accuracy}
  \subfloat[Optimal setup vs K-means]{\label{figur:Hebb_vs_kmeans}\includegraphics[width=52mm]{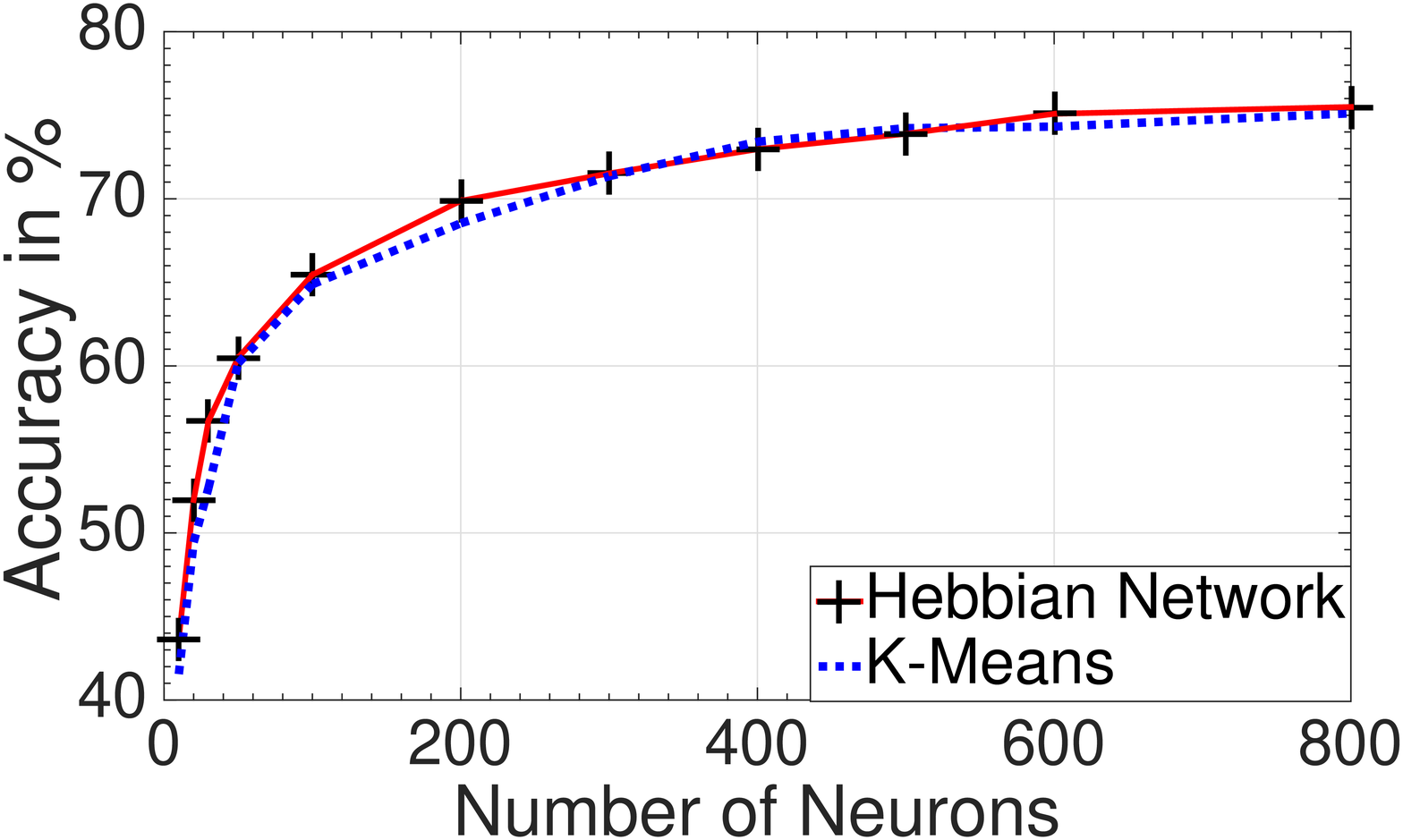}}
    \subfloat[Online training]{\label{fig:training_net_time}\includegraphics[width=52mm]{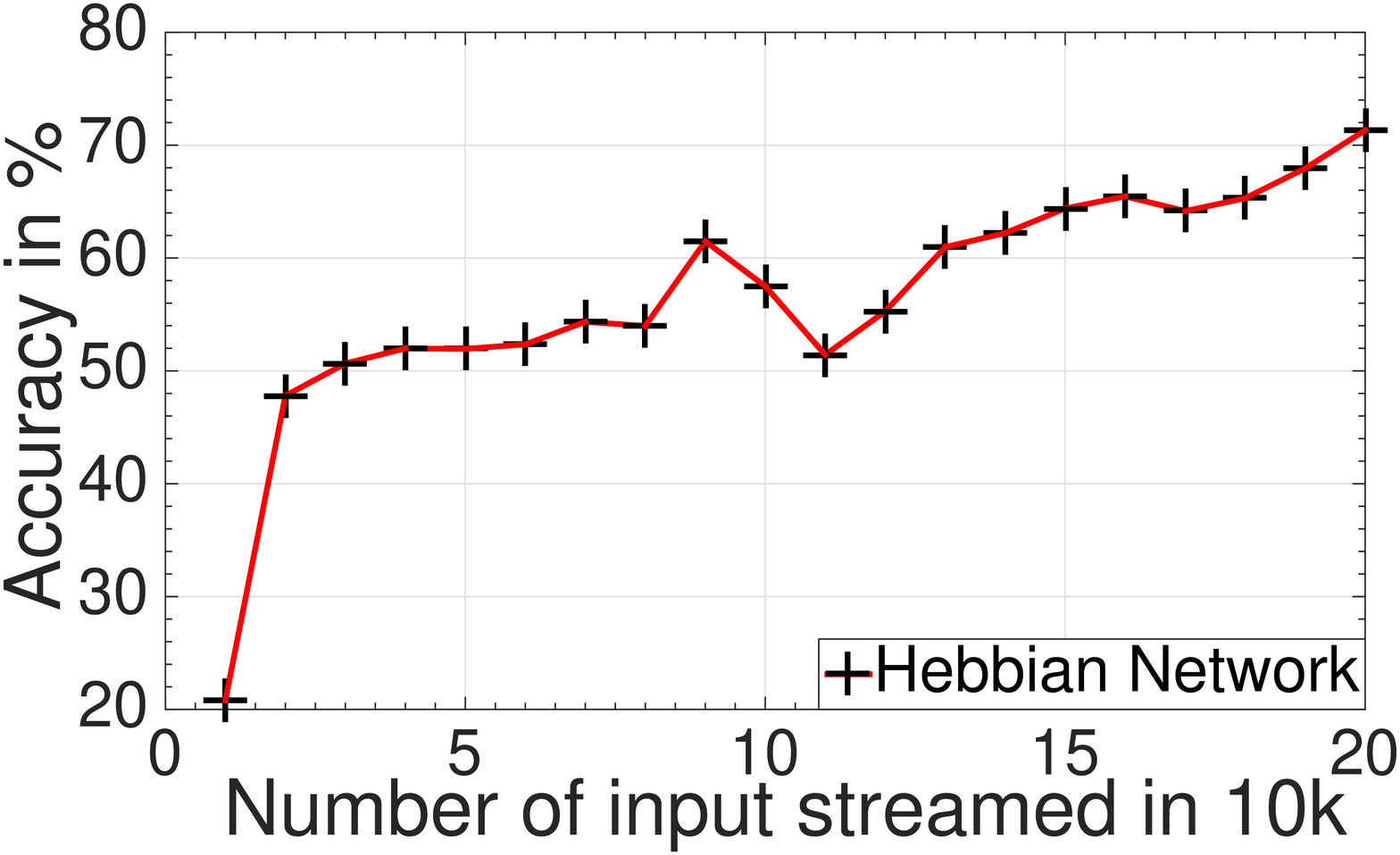}}
\end{figure}

Various unsupervised learning algorithms have been tested on the CIFAR-10 dataset. Spherical K-means, in particular, proved in \cite{coates2011analysis} to outperform autoencoders and restricted Boltzmann machines, providing a very simple and efficient solution for dictionary learning for image classification. Thus, spherical K-means is used here as a benchmark to evaluate the performance of the single-layer network. 
As with other unsupervised learning algorithms, increasing the number of output neurons to reach overcompleteness also improved classification performance (Fig.\ref{figur:Hebb_vs_kmeans}). 
Although the single-layer neural network has a higher degree of sparsity than the K-means proposed in \cite{coates2011analysis} (results not shown here), they appear to have the same performance in their optimal configurations (Fig.\ref{figur:Hebb_vs_kmeans}). 
%
% The model can be set to have varying sparsity by influencing the matrix $M^T$ in the feature extraction phase, therefore decreasing the competition between neurons. In the  case of $M^T$ being set to zero, the model Eq.\ref{activ_neurons} becomes a simple linear neuron model with ReLU activation function for which performance is well known when trained using back-propagation.

The classification accuracy of the network during training is shown in Fig.\ref{fig:training_net_time}. The graph (Fig.\ref{fig:training_net_time}) suggests that the features learned by the network over time help the system improve the classification accuracy. This is significant because it demonstrates for the first time the effectiveness of features learned with a Hebb-like cost-function minimisation.
It is not obvious a priori that the online optimisation of a cost-function for sparse similarity matching (Eq.\ref{eq:local}) produces features suitable for image classification.
% \vspace{-5mm}

As shown in Table \ref{Tab:Multi_resolution}, the multi-resolution network outperforms the single resolution network and K-means algorithm \cite{coates2011analysis}, reaching 80.42\% accuracy on the CIFAR-10. The multi-resolution model shows better performance, while requiring less computation and memory than the single resolution model. It also outperforms the single layer NOMP \cite{lin2014stable}, sparse TIRBM \cite{sohn2012learning}, CKN-GM and CKN-PM \cite{mairal2014convolutional}, which are more complex models. It was outperformed only by combined models or models with three layers or more. 
%\vspace{-1mm}
\newpage
\begin{table}[h!]
  \centering
\begin{tabular}{ |p{9cm}|p{1.5cm}|  }
\hline
Algorithm & Accuracy \\
\hline
\hline
\textbf{Single-Layer, Single Resolution (4k neurons)}   & \textbf{79.58 \%} \\
\textbf{Single-Layer, Multi-Resolution (3$\times$1.6k neurons)}   & \textbf{80.42 \%} \\
\hline
% \textbf{Multi-Layer, Single Resolution (200+800 neurons)}   & \textbf{75.13 \%}   \\
% \hline
\hline
Single-layer K-means \cite{coates2011analysis} (4k neurons) & 79.60 \%  \\
Multi-layer K-means \cite{coates2011analysis} (3 Layers, $>$4k neurons) & 82.60 \%  \\
\hline
Sparse RBM & 72.40  \%  \\
Convolutional DBN \cite{krizhevsky2010convolutional} & 78.90 \%  \\
Sparse TIRBM \cite{sohn2012learning} (4k neurons)  & 80.10\%  \\
TIOMP-1/T \cite{sohn2012learning} (combined transformations, 4k neurons) & 82.20 \% \\
\hline
Single Layer NOMP \cite{lin2014stable} ( 5k neurons)  & 78.00 \%  \\
Multi-Layer NOMP \cite{lin2014stable}  (3 Layers, $>$4k neurons) & 82.90 \%  \\
\hline
Multi-Layer CKN-GM  \cite{mairal2014convolutional} & 74.84  \%  \\
Multi-Layer  CKN-PM \cite{mairal2014convolutional}  & 78.30 \%  \\
Multi-Layer CKN-CO \cite{mairal2014convolutional} (combining CKN-GM \& CKN-PM)  & 82.18 \%  \\
\hline
\end{tabular}
  \caption{ Comparison of the single-layer network with unsupervised learning algorithms on CIFAR-10.}\label{Tab:Multi_resolution}
\end{table}
\vspace{-12mm}
\subsection{Evaluation of the Multi-layer Model}

%Learning overcomplete representations at every stage of a multi-layer network is a challenging task due to the increasing number of neurons required. 
A single resolution, double-layer neural network with different numbers of neurons in each layer was trained similarly to the single-layer network in the previous section.
In Table \ref{Tab:Second_layer}, $\phi_1$ and $\phi_2$ correspond respectively to the features learned by the first and second layer. The results show that $\phi_2$ alone are less discriminative than $\phi_1$ as indicated in Fig. \ref{figur:Hebb_vs_kmeans}. % for the same number of neurons. 
However, when combined ($\phi_1+\phi_2$) the model achieves better performance than each layer considered separately. 
%
%Thus it seems that the combined features become beneficial when the second layer is much larger than the first layer. 
%
Nevertheless, the preliminary results indicate that the sizes of the two layers unevenly affect the performance of the network. A future test may investigate if a multi-layer architecture can outperform the largest shallow networks. 
%
% A future test may analyse whether the second layer learns larger-scale features than the first layer, a finding that could explain the improved classification accuracy with the multi-layer neural network. %Such test includes investigating the adaptation of the multi-resolution to multi-layer architectures. 
%
\begin{table}[!ht]
  \centering
\begin{tabular}{cc|c|c|c|c|c|}
\cline{3-7}
& & \multicolumn{5}{ c| }{\#Neurons Layer 2} \\ \cline{3-7}
& & 50 & 100 & 200 & 400 & 800 \\ \cline{1-7}
\multicolumn{1}{ |c| }{\multirow{2}{*}{100 Neurons Layer 1} } &
\multicolumn{1}{ |c| }{$\phi_2$} & 54.9\% & 59.7\% & 64.7\% & 68.7\% &   71.45\%  \\ \cline{2-7}
\multicolumn{1}{ |c| }{}                        &
\multicolumn{1}{ |c| }{$\phi_1$+$\phi_2$} & 67.2\% & 68.1\% & 69.9\% & 72.4\% &   73.81\%   \\ \cline{1-7}
\multicolumn{1}{ |c|  }{\multirow{2}{*}{200 Neurons Layer 1} } &
\multicolumn{1}{ |c| }{$\phi_2$} & 55.8\% & 60.6\% & 65.3\% & 70.3\% &   72.7\% \\ \cline{2-7}
\multicolumn{1}{ |c|  }{}                        &
\multicolumn{1}{ |c| }{$\phi_1$+$\phi_2$} & 69.9\% & 70.8\% & 71.9\% & 73.7\% &   75.1\%  \\ \cline{1-7}
\end{tabular}
\caption{Classification accuracy for a two-layer network.}\label{Tab:Second_layer}
\end{table}
%
% ****************************************************************************
% ****************************************************************************
% ****************************************************************************
% ****************************************************************************
% ****************************************************************************
% Conclusion AREA
% ****************************************************************************
% ****************************************************************************
% ****************************************************************************
% ****************************************************************************
% ****************************************************************************
%
\vspace{-12mm}
\section{Conclusion}
%\vspace{-4mm}
%
This work proposes a multi-layer neural network exploiting Hebbian/anti-Hebbian rules to learn features for image classification. The network is trained on the CIFAR-10 image dataset prior to feeding a linear classifier. 
The model successfully learns online more discriminative representations of the data when the number of neurons and the number of layers increase. %We observed in this work that
The overcompleteness of the representation is critical for learning relevant features. %We have also shown 
The results show that a minimum unsupervised learning time is needed to optimise the network leading to better classification accuracy. 
Finally, one key factor in improving image classification is the appropriate choice of the receptive field size used for training the network.

Such findings prove that neural networks can be trained to solve problems as complex as sparse dictionary learning with Hebbian learning rules, delivering competitive accuracy compared to other encoder, including deep neural networks. This makes deep Hebbian networks attractive for building large-scale image classification systems. 
The competitive performances on the CIFAR-10 suggests that this model can offer an alternative to batch trained neural networks. Ultimately, thanks to its bio-inspired architecture and learning rules, it also stands as a good candidate for memristive devices \cite{poikonen2016online}. 
Moreover, if a decaying factor is added to the proposed model that might result in an algorithm that can deal with complex datasets with temporal variations of the distributions.

% ****************************************************************************
% BIBLIOGRAPHY AREA
% ****************************************************************************

\vspace{-2mm}

% EITHER use the included BST file
 \bibliographystyle{splncs03}
 \bibliography{bib_esann_v2}

\end{document}